\documentclass{article}

\usepackage{arxiv}

\usepackage[utf8]{inputenc} 
\usepackage[T1]{fontenc}    
\usepackage{hyperref}       
\usepackage{url}            
\usepackage{booktabs}       
\usepackage{amsfonts}       
\usepackage{nicefrac}       
\usepackage{microtype}      
\usepackage{lipsum}

\usepackage{bm}
\usepackage{amsmath}
\usepackage{amssymb}
\usepackage{amsfonts}
\usepackage{graphicx}
\def\vec#1{\mathbf{#1}}
\usepackage{algorithm}
\usepackage{algorithmic}
\DeclareMathOperator*{\argmax}{arg\,max}

\title{Training Deep Models to be Explained with Fewer Examples}

\author{
  Tomoharu Iwata\\
  NTT Communication Science Laboratories\\
  \AND
  Yuya Yoshikawa\\
  Software Technology and Artificial Intelligence Research Laboratory\\
  Chiba Institute of Technology\\
}
\date{}

\begin{document}
\maketitle

\begin{abstract}
  Although deep models achieve high predictive performance, it is difficult for humans to understand the predictions they made. Explainability is important for real-world applications to justify their reliability.  Many example-based explanation methods have been proposed, such as representer point selection, where an explanation model defined by a set of training examples is used for explaining a prediction model. For improving the interpretability, reducing the number of examples in the explanation model is important. However, the explanations with fewer examples can be unfaithful since it is difficult to approximate prediction models well by such example-based explanation models. The unfaithful explanations mean that the predictions by the explainable model are different from those by the prediction model. We propose a method for training deep models such that their predictions are faithfully explained by explanation models with a small number of examples. We train the prediction and explanation models simultaneously with a sparse regularizer for reducing the number of examples. The proposed method can be incorporated into any neural network-based prediction models. Experiments using several datasets demonstrate that the proposed method improves faithfulness while keeping the predictive performance.
\end{abstract}

\section{Introduction}

Explainability is important for real-world applications of machine learning models
especially in critical decision-making domains, 
such as medical diagnosis, loan decision,
and personnel evaluation~\cite{adadi2018peeking,du2019techniques,tjoa2020survey,arrieta2020explainable}.
For users,
explanations help to understand a specific prediction given by the model,
improve the trustworthiness,
and check unintended unfair decision-making.
For researchers and developers,
explanations help to understand, debug and improve the model,
and discover knowledge in the data.

Due to the development of deep learning, many complicated models have been proposed.
Although these models can achieve high predictive performance,
their predictions are unexplainable.
For explaining predictions by deep models,
many methods have been proposed~\cite{ribeiro2016should,NIPS2017_8a20a862,ribeiro2018anchors,selvaraju2017grad,koh2017understanding,kim2018interpretability,yeh2018representer}.

Representer Point Selection (RPS)~\cite{yeh2018representer}
is a representative method for explaining deep models by examples.
With RPS, predictions are modeled by
a linear combination of activations of training examples,
where the linear weight corresponds to
the importance of the training example for the predictions.
By fitting the explanation model to a given pretrained prediction model,
we can understand the prediction model.
To improve the interpretability,
it is important to reduce the number of examples used
in the explanation model.
For humans, explanations with many examples
are difficult to understand.
However, it is difficult to approximate the prediction model well
by the explanation model with a small number of examples,
and the explanations can be unfaithful.
``Unfaithful'' means that the approximated models for explanations
do not precisely reflect the behavior of the original model~\cite{du2019techniques}.

In this paper,
we propose a method for training deep models
such that the prediction model achieves high predictive
performance and the explanation model achieves high faithfulness with
a small number of examples.
We train the prediction and explanation models simultaneously with a sparse regularizer for reducing the number of examples. 
The proposed method is applicable to any neural network-based models.
In general, there is a trade-off between predictive performance and interpretability.
By changing the weight of our regularizer in the loss function,
we can control the trade-off depending on applications.

Our major contributions are as follows:
\begin{enumerate}
\item We propose a framework to train models such that the faithfulness of example-based explanations is improved with fewer examples.
\item We incorporate the stochastic gates to obtain sparse explanations.
\item We experimentally confirm that the proposed method improves the faithfulness while keeping the predictive performance.
\end{enumerate}
The remainder of this paper is organized as follows.
In Section~\ref{sec:related}, we briefly describe related work.
In Section~\ref{sec:proposed}, we formulate our problem,
propose our method for training neural networks that are explained with a small number of examples,
and present its training procedure.
In Section~\ref{sec:experiments}, we demonstrate that the effectiveness of the proposed method.
Finally, we present concluding remarks and discuss future work in Section~\ref{sec:conclusion}.

\section{Related work}
\label{sec:related}

For the explanation of deep models, there are two main approaches: feature-based, and example-based.
With the feature-based approach,
such as local interpretable model-agnostic explanations (LIME)~\cite{ribeiro2016should},
SHapley Additive exPlanations (SHAP)~\cite{NIPS2017_8a20a862},
and Anchors~\cite{ribeiro2018anchors},
features that are important for the prediction are extracted.
With the example-based approach,
training examples that are important for the prediction are extracted.
We focus on the example-based approach in this paper.

There have been proposed a number of example-based method~\cite{bien2011prototype,kim2014bayesian,kim2016examples,koh2017understanding,anirudh2017influential,yeh2018representer}.
For example, influence functions are used for characterizing the influence
of each training example in terms of change in the loss~\cite{koh2017understanding}.
The proposed method is based on RPS~\cite{yeh2018representer}
since RPS considers both the positive and negative influence of training examples,
and is scalable.
RPS fails when the given pretrained model is unexplainable,
or the given model is difficult to be approximated by the explanation models.
The proposed method makes the model explainable by RPS.

Another approach for interpretable machine learning is to use
explainable models that are easy to interpret, such as
linear models, decision trees, and rule-based models.
However, the predictive performance of such simple models would be low.
To achieve both high performance and high interpretability,
neural network-based methods have been proposed~\cite{schwab2019granger,yoshikawa2020neural,sabour2017dynamic,zhou2016learning,selvaraju2017grad,yoshikawa2020gaussian}.
However, they need to change the model architecture,
and additional model parameters to be trained are need. 
In contrast, the proposed method can make any models explainable.
The interpretable convolutional neural network~\cite{zhang2018interpretable}
is a regularizer-based method as with the proposed method.
However, it is specific convolutional neural networks (CNNs).

Explanation-based optimization (EXPO)~\cite{plumb2020regularizing}
is related to the proposed method since it trains prediction models such that
the explanations become faithful.
However, EXPO is a feature-based explanation approach,
and it does not consider reducing the number of examples for the explanation.

\section{Proposed method}
\label{sec:proposed}

\subsection{Problem formulation}

Suppose that we are given training data $\mathcal{D}=\{(\vec{x}_{n},y_{n})\}_{n=1}^{N}$,
where $\vec{x}_{n}$ is the feature vector of the $n$th example,
$y_{n}$ is its label,
and $N$ is the number of training examples.
For classification tasks, $y_{n}\in\{0,1\}^{C}$ is a one-hot vector that represents its class label,
and for regression tasks, $y_{n}\in\mathbb{R}^{C}$.
Our aim is to learn prediction model $\hat{y}=g(\vec{x})$
that achieves high test predictive performance,
and
that is explained by fewer training examples with high faithfulness.

\subsection{Model}

\begin{figure}[t!]
\centering
\includegraphics[width=18em]{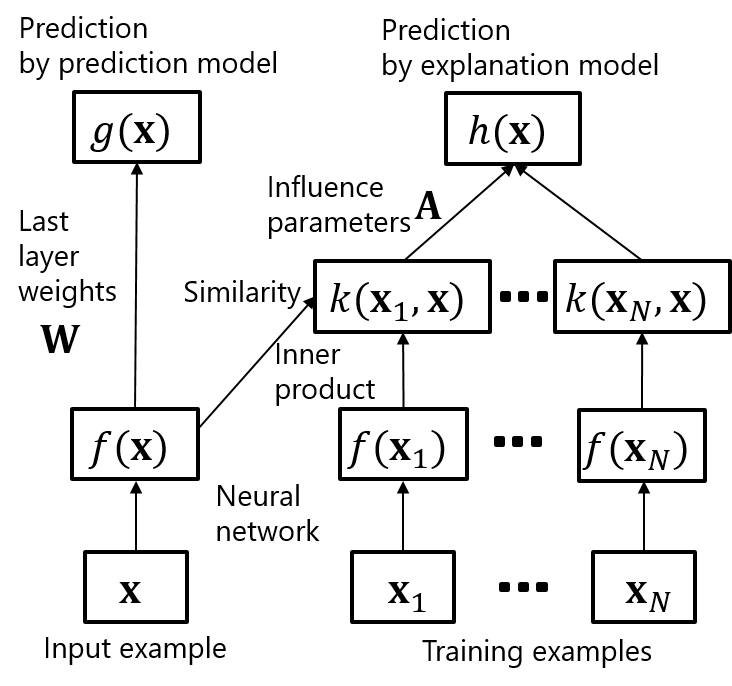}
\caption{Our model architecture. The prediction and explanation models share neural network $f$. The prediction model outputs the prediction using linear weights $\vec{W}$ as defined in Eq.~(\ref{eq:g}). The explanation model outputs the prediction using the similarity $k$ in Eq.~(\ref{eq:k}) and influence parameters $\vec{A}$ as defined in Eq.~(\ref{eq:h}).}
\label{fig:model}
\end{figure}

We consider a neural network as a prediction model, which can be written as follows,
\begin{align}
g(\vec{x};\vec{W},\bm{\Theta})= r(\vec{W}^{\top}f(\vec{x};\bm{\Theta})),
\label{eq:g}
\end{align}
where 
$f:\mathbb{R}^{D}\rightarrow\mathbb{R}^{K}$ is a neural network before the last layer,
$\bm{\Theta}$ is its parameters,
$\vec{W}\in\mathbb{R}^{K\times C}$ is a linear projection matrix,
$K$ is the number of units of the last layer,
$r$ is the softmax function for classification tasks,
and $r$ is the identity function for regression tasks.

With a representer theorem~\cite{yeh2018representer},
we can approximate prediction model $g(\vec{x};\vec{W},\bm{\Theta})$
using training examples $\{\vec{x}_{n}\}_{n=1}^{N}$,
\begin{align}
g(\vec{x};\vec{W},\bm{\Theta}) \approx h(\vec{x};\vec{A},\bm{\Theta})
=r\left(\sum_{n=1}^{N}\bm{\alpha}_{n}f(\vec{x}_{n};\bm{\Theta})^{\top}f(\vec{x};\bm{\Theta})\right),
\label{eq:h}
\end{align}
where $\bm{\alpha}_{n}\in\mathbb{R}^{C}$, 
$\vec{A}=\{\bm{\alpha}_{n}\}_{n=1}^{N}$,
and $h(\vec{x};\vec{A},\bm{\Theta})$ is the explanation model.
The inner product
\begin{align}
k(\vec{x}_{n},\vec{x};\bm{\Theta})=f(\vec{x}_{n};\bm{\Theta})^{\top}f(\vec{x};\bm{\Theta}),
\label{eq:k}
\end{align}
represents
the similarity between training example $\vec{x}_{n}$ and example to be predicted $\vec{x}$.
Parameter $\alpha_{nc}$ represents the influence of the $n$th training example
on predicting class $c$, where $\alpha_{nc}>0$ ($\alpha_{nc}<0$) indicates that
the probability of class $c$ increases (decreases) when
the example that is similar to the $n$th training example,
and $\alpha_{nc}=0$ indicates that the $n$th training example has no influence on the prediction.
Therefore, $\alpha_{nc}$ is used for explaining the prediction model.
Figure~\ref{fig:model} shows the proposed model.

By checking the outputs of the prediction and explanation models,
we can judge whether the explanation by the explanation model
is faithful or not for each data point without its true label information.
When their outputs are the same,
\begin{align}
\argmax_{c}g_c(\vec{x};\vec{W},\bm{\Theta})=\argmax_{c}h_c(\vec{x};\vec{A},\bm{\Theta}),
\end{align}
it is faithful,
where $g_c$ is the $c$th output unit of neural network $g$.
When their outputs are different,
\begin{align}
\argmax_{c}g_c(\vec{x};\vec{W},\bm{\Theta})\neq\argmax_{c}h_c(\vec{x};\vec{A},\bm{\Theta}),
\end{align}
it is unfaithful.

When there are many nonzero $\alpha_{nc}$, it is difficult for humans
to understand the explanations.
For improving the interpretability,
fewer nonzero $\alpha_{nc}$ is preferable.
However, with a small number of $\alpha_{nc}$,
it is difficult to approximate prediction model $g$ well
with explanation model $h$ in Eq.~(\ref{eq:h}).
Then, we train prediction model $g$ and explanation model $h$ such that
$g$ is approximated well by $h$ with a smaller number of nonzero elements of $\vec{A}$
while keeping the predictive performance of $g$
by minimizing the following objective function,
\begin{align}
E(\bm{\Theta},\vec{W},\vec{A})
= \mathbb{E}_{(\vec{x},y)}[\ell(y,g(\vec{x};\vec{W},\bm{\Theta}))]
+ 
\lambda\mathbb{E}_{\vec{x}}[\ell(g(\vec{x};\vec{W},\bm{\Theta}),h(\vec{x};\vec{A},\bm{\Theta}))]
+\eta \Omega(\vec{A}),
\end{align}
where $\mathbb{E}$ represents the expectation,
$\ell$ is the loss function,
$\Omega(\vec{A})$ is a regularizer for sparsity,
and $\lambda>0$ and $\eta>0$ are hyperparameters.
The first term tries to improve the prediction performance by minimizing the loss between
the training data and predction by $g$.
The second term tries to improve the approximation of $g$ by $h$ by minimizing the loss between
the prediction by $g$ and that by $h$.
The third term tries to decrease the number of nonzero elements of $\vec{A}$.

For the sparse regularization, we use stochastic gates~\cite{yamada2020feature}.
With the stochastic gates,
sparse $\vec{A}$ is parameterized using continuous relaxation
of Bernoulli variables $\vec{Z}\in [0,1]^{N\times C}$,
\begin{align}
\vec{A}=\vec{Z}\odot\vec{S},
\label{eq:A}
\end{align}
where $\vec{S}\in\mathbb{R}^{N\times C}$, and $\odot$ is the elementwise multiplication.
The relaxed Bernoulli variable is
\begin{align}
z_{nc}=\max(0,\min(1,\mu_{nc}+\epsilon_{nc})),
\label{eq:z}
\end{align}
where $\mu_{nc}$ is a parameter, and $\epsilon_{nc}$ is drawn from Gaussian distribution $\mathcal{N}(0,\sigma^{2})$.
The regularization term is the sum of the probabilities that gates
$\vec{Z}=\{\{z_{nc}\}_{c=1}^{C}\}_{n=1}^{N}$ are active, or
$\sum_{n=1}^{N}\sum_{c=1}^{C}p(z_{nc}>0)$, and then,
\begin{align}
\mathbb{E}_{\vec{Z}|\vec{M}}[\Omega(\vec{Z})]=\sum_{n=1}^{N}\sum_{c=1}^{C}\Phi\left(\frac{\mu_{nc}}{\sigma}\right),
\label{eq:omega}
\end{align}
where $\Phi$ is the standard Gaussian cumulative density function,
and $\vec{M}=\{\{\mu_{nc}\}_{c=1}^{C}\}_{n=1}^{N}$.
The objective function with the stochastic gates is written by
\begin{align}
E(\bm{\Theta},\vec{W},\vec{S},\vec{M})
= \mathbb{E}_{(\vec{x},y)}[\ell(y,g(\vec{x};\vec{W},\bm{\Theta}))]
+
\lambda\mathbb{E}_{\vec{x}}[\mathbb{E}_{\vec{Z}|\vec{M}}[\ell(g(\vec{x};\vec{W},\bm{\Theta}),h(\vec{x};\vec{S},\vec{Z},\bm{\Theta}))]]
+\eta\mathbb{E}_{\vec{Z}|\vec{M}}[\Omega(\vec{Z})],
\label{eq:E}
\end{align}
where parameters $\vec{A}$ are replaced by $\vec{S}$ and $\vec{M}$.

\subsection{Training}

Algorithm~\ref{alg} shows the training procedure of the proposed model
based on a stochastic gradient descent method.
The parameters to be estimated are $\bm{\Theta},\vec{W},\vec{S}$, and $\vec{M}$.
The expectation over relaxed Bernoulli variables $\vec{Z}$
in the second term in Eq.~(\ref{eq:E}) is calculated
by Monte Calro sampling.
Then, the objective function is 
\begin{align}
E(\bm{\Theta},\vec{W},\vec{S},\vec{M})
&= \frac{1}{B}\sum_{(\vec{x},y)\in\mathcal{B}}[\ell(y,g(\vec{x};\vec{W},\bm{\Theta}))]
+
\frac{\lambda}{BJ}\sum_{\vec{x}\in\mathcal{B}}\sum_{j=1}^{J}[\ell(g(\vec{x};\vec{W},\bm{\Theta}),h(\vec{x};\vec{S},\vec{Z}^{(j)},\bm{\Theta}))]
\nonumber\\
&+\eta\sum_{n=1}^{N}\sum_{c=1}^{C}\Phi\left(\frac{\mu_{nc}}{\sigma}\right),
\label{eq:Eminibatch}
\end{align}
where $\mathcal{B}$ is the minibatch of randomly sampled training example,
$B$ is its size, and $\vec{Z}^{(j)}$ is the $j$th samples of relaxed Bernoulli variable $\vec{Z}$,
which are obtained by Eq.~(\ref{eq:z}).

\begin{algorithm}[t]
  \caption{Training procedure of the proposed model.}
  \label{alg}
  \begin{algorithmic}[1]
    \renewcommand{\algorithmicrequire}{\textbf{Input:}}
    \renewcommand{\algorithmicensure}{\textbf{Output:}}
    \REQUIRE{Training data $\mathcal{D}=\{(\vec{x}_{n},y_{n})\}_{n=1}^{N}$.}
    \ENSURE{Trained model parameters $\bm{\Theta},\vec{W},\vec{S},\vec{M}$}
    \STATE Initialize parameters $\bm{\Theta},\vec{W},\vec{S},\vec{M}$.
    \WHILE{not done}
    \STATE Randomly sample minibatch $\mathcal{B}$ from training data $\mathcal{D}$.
    \FOR{$j=1,\cdots,J$}
    \FOR{$n=1,\cdots,N$}
    \FOR{$c=1,\cdots,C$}
    \STATE Randomly sample $\epsilon_{nc}^{(j)}$ from the Gaussian distribution with mean zero and variance $\sigma^{2}$.
    \STATE Calculate $z_{nc}^{(j)}$ by Eq.~(\ref{eq:z}).
    \ENDFOR
    \ENDFOR
    \ENDFOR
    \STATE Calculate minibatch loss $E$
    by Eq.~(\ref{eq:Eminibatch}) and its gradient.
    \STATE Update parameters $\bm{\Theta}, \vec{W}, \vec{S}, \vec{M}$ by a stochastic gradient method using the gradient.
    \ENDWHILE
  \end{algorithmic}
\end{algorithm}

\section{Experiments}
\label{sec:experiments}

\subsection{Data}

We evaluated the proposed method using four datasets: Glass, Vehicle, Segment, and CIFAR10.
Glass consists of 214 examples represented by
nine types of oxide content from six classes of glasses.
Vehicle consists of 946 silhouettes of the objects represented by 18 shape features
from four types of vehicles.
Segment consists of 2,310 images described by nine high-level numeric-valued attributes from seven classes.
Glass, Vehicle, and Segment data were obtained from LIBSVM~\cite{chang2011libsvm}.
CIFAR10~\cite{krizhevsky2009learning} consists of images from 10 classes.
We randomly sampled 1,000 images, and extracted 1,000-dimensional feature vectors using
deep residual network~\cite{he2016deep} with 18 layers as preprocessing.
For each dataset,
we used 70\% of the data points for training,
10\% for validation, and the remainder for testing.
We performed 10 experiments with different train, validation, and test splits,
and evaluated their average.

\subsection{Setting}

For the model, we used four-layered feed-forward
neural networks.
We used hidden layers with 512 units for CIFAR10 data, and 64 units for the other data.
The activation function in the neural networks was rectified linear unit, $\mathrm{ReLU}(x)=\max(0,x)$.
Optimization was performed using Adam~\cite{kingma2014adam}
with learning rate $10^{-3}$,
dropout ratio 0.3~\cite{srivastava2014dropout},
and batch size 32.
We used $\lambda=1$, $J=1$, and $\sigma=0.5$.
$\eta$ was tuned from $\{0,10^{-4},10^{-3},10^{-2},10^{-1},1\}$ using the validation data,
where the geometric mean of the test accuracy and faithfulness was used as the measurement for the tuning.
The maximum number of training epochs was 10,000,
and the validation data were used for early stopping.
We implemented the proposed method with PyTorch~\cite{paszke2017automatic}.

\subsection{Comparing methods}

We compared the proposed method (Ours) with the following five methods:
Joint, RPS, RPSR, Pretrain, and PretrainR. 
The Joint method jointly trains prediction model $g$ and explanation model $h$
without the sparse regularization as follows,
\begin{align}
E(\bm{\Theta},\vec{W},\vec{A})
= \mathbb{E}_{(\vec{x},y)}[\ell(y,g(\vec{x};\vec{W},\bm{\Theta}))]
+
\lambda\mathbb{E}_{\vec{x}}[\ell(g(\vec{x};\vec{W},\bm{\Theta}),h(\vec{x};\vec{A},\bm{\Theta}))].
\end{align}
The Joint method corresponds to the proposed method without the regularization term.
The RPS method is the representer point selection method~\cite{yeh2018representer}
that uses explanation model $h$ for prediction,
where explanation model $h$ is trained by minimizing
the loss between true labels and the predictions
by explanation model $h$,
\begin{align}
E(\bm{\Theta},\vec{A})
= \mathbb{E}_{(\vec{x},y)}[\ell(y,h(\vec{x};\vec{A},\bm{\Theta}))].
\end{align}
The RPSR method is the RPS method with the regularization,
where the objective function is
\begin{align}
E(\bm{\Theta},\vec{S},\vec{M})
= \mathbb{E}_{\vec{x}}[\mathbb{E}_{\vec{Z}|\vec{M}}[\ell(y,h(\vec{x};\vec{S},\vec{Z},\bm{\Theta}))]]
+\eta\mathbb{E}_{\vec{Z}|\vec{M}}[\Omega(\vec{Z})].
\end{align}
With the Pretrain method, prediction model $g$ is pretrained by minimizing
the loss between the true and predicted labels,
\begin{align}
E(\bm{\Theta},\vec{W})
= \mathbb{E}_{(\vec{x},y)}[\ell(y,g(\vec{x};\vec{W},\bm{\Theta}))].
\end{align}
Then, explanation model $h$ is trained while fixing $f$ by minimizing
the loss between the predicted labels by prediction model $g$ and explanation model $h$,
\begin{align}
E(\vec{A})
= \mathbb{E}_{\vec{x}}[\ell(g(\vec{x};\vec{W},\bm{\Theta}),h(\vec{x};\vec{A},\bm{\Theta}))].
\end{align}
The PretrainR method is the Pretrain method with the regularization,
where explanation model $h$ is trained by minimizing the following objective function
after the pretraining of prediction model $g$,
\begin{align}
E(\vec{S},\vec{M})
= \mathbb{E}_{\vec{x}}[\mathbb{E}_{\vec{Z}|\vec{M}}[\ell(g(\vec{x};\vec{W},\bm{\Theta}),h(\vec{x};\vec{S},\vec{Z},\bm{\Theta}))]]
+\eta\mathbb{E}_{\vec{Z}|\vec{M}}[\Omega(\vec{Z})].
\end{align}

\subsection{Evaluation measurement}

For evaluating the predictive performance of the model,
we used the test accuracy (higher is better)
that represents the discrepancy between the true labels and the prediction by neural network $g$
for the test data,
\begin{align}
  \text{Accuracy}
  =\frac{1}{|\mathcal{D}_{\text{T}}|}\sum_{(\vec{x},y)\in\mathcal{D}_{\text{T}}}I\left(y=\argmax_{c}g_c(\vec{x};\vec{W},\bm{\Theta})\right),
\end{align}
where $\mathcal{D}_{\text{T}}$ is the test data,
$|\mathcal{D}_{\text{T}}|$ is its size,
and $I(\cdot)$ is the indicator function, i.e., $I(A)=1$ if $A$ is true, and $I(A)=0$ otherwise.
For evaluating faithfulness of the explanations,
we used the faithfulness score (higher is better)
that represents the discrepancy between prediction model $g$ and explanation model $h$ for the test data,
\begin{align}
  \text{Faithfulness}=\frac{1}{|\mathcal{D}_{\text{T}}|}\sum_{\vec{x}\in\mathcal{D}_{\text{T}}}I\left(\argmax_{c}g_c(\vec{x};\vec{W},\bm{\Theta})
 =\argmax_{c}h_c(\vec{x};\vec{S},\vec{M},\bm{\Theta})\right).
\end{align}
For selecting a given number of training examples used for explanations,
we used the expectation of the influence parameter,
\begin{align}
\mathbb{E}_{z_{nc}|\mu_{nc}}[\alpha_{nc}]=\max(0,\min(1,\mu_{nc})),
\end{align}
in methods that use the stochastic gates,
i.e., the proposed, RPSR, and PretrainR methods.
The explanation model with $U$ training examples for explanations
is modeled by
\begin{align}
h(\vec{x};\vec{S},\vec{M},\bm{\Theta})
=r\left(\sum_{n=1}^{N}\sum_{c=1}^{C}\tilde{\alpha}_{nc}f(\vec{x}_{n};\bm{\Theta})^{\top}f(\vec{x};\bm{\Theta})\right),
\end{align}
where
\begin{align}
\tilde{\alpha}_{nc}=\left\{
\begin{array}{ll}
\mathbb{E}_{z_{nc}|\mu_{nc}}[\alpha_{nc}] & 
\text{if $\mathbb{E}_{z_{nc}|\mu_{nc}}[\alpha_{nc}]$ is in the top $U$ among $n=1,\dots,N$, $c=1,\dots,C$}\\
0 & \text{otherwise.}
\end{array}
\right.
\label{eq:alphatilde}
\end{align}
In methods that do not use the stochastic gates, i.e., the Joint, RPS, and Pretrain methods,
$\mathbb{E}_{z_{nc}|\mu_{nc}}[\alpha_{nc}]$ was replaced by $\alpha_{nc}$
in Eq.~(\ref{eq:alphatilde}).

\subsection{Results}

\begin{figure*}[t!]
\centering
{\tabcolsep=3em
\begin{tabular}{cc}
\includegraphics[width=15em]{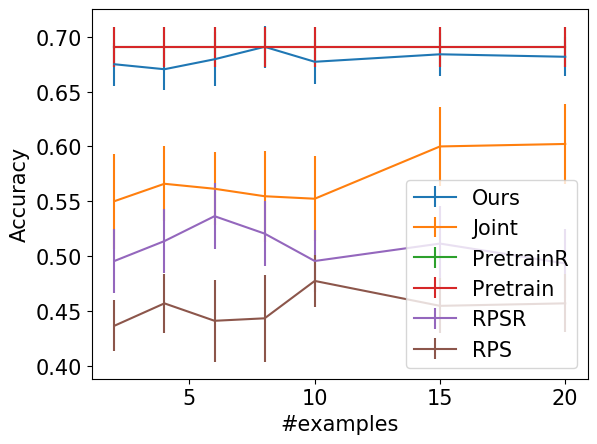}&
\includegraphics[width=15em]{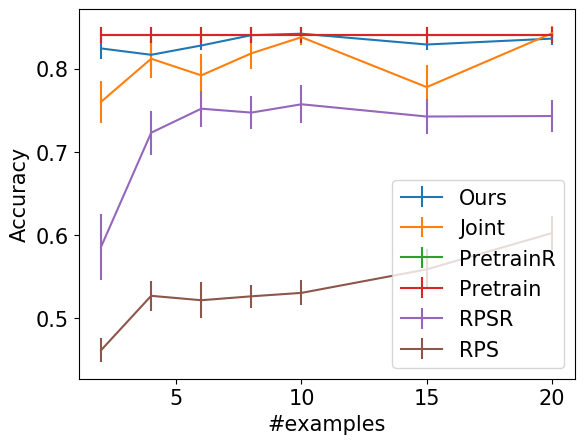}\\
(a) Glass & (b) Vehicle \\
\includegraphics[width=15em]{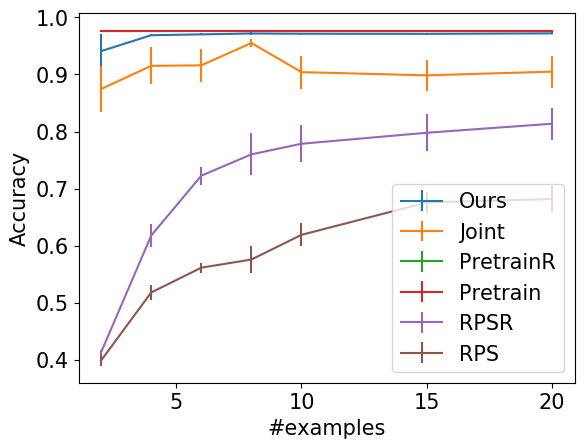}&
\includegraphics[width=15em]{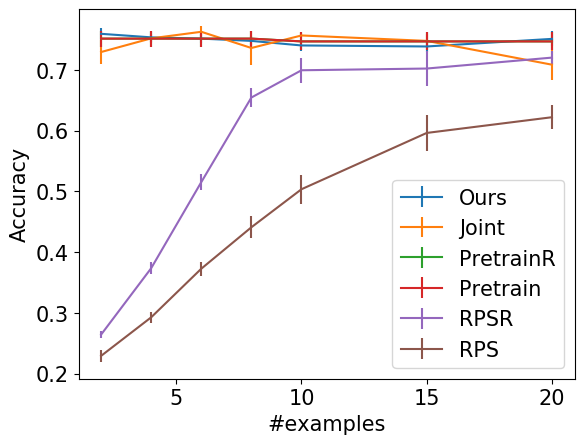}\\
(c) Segment & (d) CIFAR10 \\
\end{tabular}}
\caption{Test accuracy with different numbers of training examples used for explanations $U$. The bar shows the standard error.}
\label{fig:accuracy}
\end{figure*}

\begin{figure*}[t!]
\centering
{\tabcolsep=3em
\begin{tabular}{cc}
\includegraphics[width=15em]{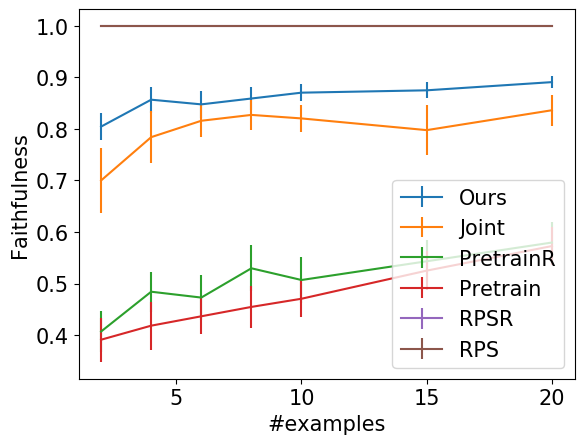}&
\includegraphics[width=15em]{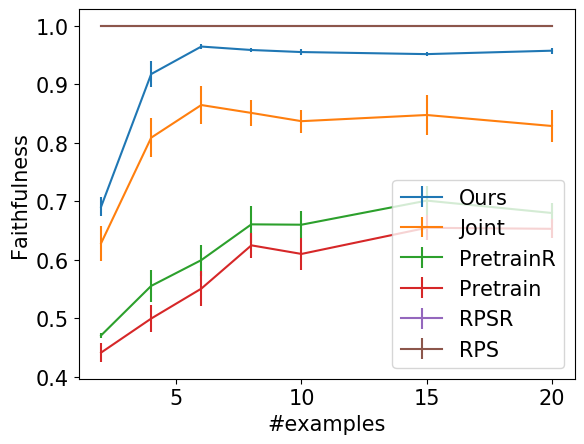}\\
(a) Glass & (b) Vehicle \\
\includegraphics[width=15em]{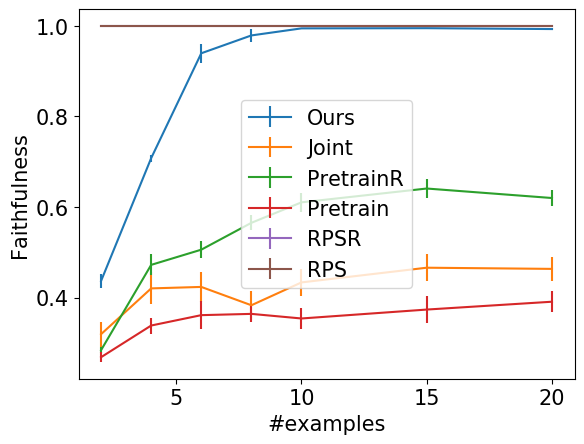}&
\includegraphics[width=15em]{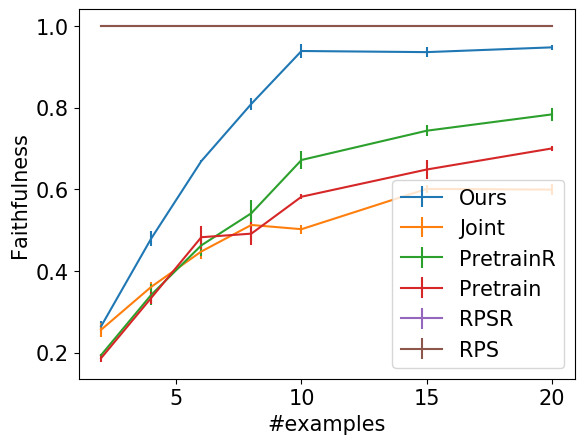}\\
(c) Segment & (d) CIFAR10 \\
\end{tabular}}
\caption{Test faithfulness with different numbers of training examples used for explanations $U$. The bar shows the standard error.}
\label{fig:faithfulness}
\end{figure*}

\begin{figure*}[t!]
\centering
{\tabcolsep=1em
\begin{tabular}{ccccc}
\includegraphics[width=4.3em]{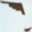} &
\includegraphics[width=4.3em]{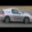} &
\includegraphics[width=4.3em]{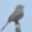} &
\includegraphics[width=4.3em]{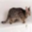} &   
\includegraphics[width=4.3em]{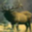} \\
(a) Airplain & (b) Automobile & (c) Bird & (d) Cat & (e) Deer \\
\includegraphics[width=4.3em]{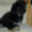} &     
\includegraphics[width=4.3em]{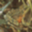} &     
\includegraphics[width=4.3em]{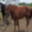} &     
\includegraphics[width=4.3em]{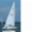} &   
\includegraphics[width=4.3em]{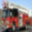} \\
(f) Dog & (g) Frog & (h) Horse & (i) Ship & (j) Truck \\
\end{tabular}}
\caption{Ten examples used for the explanations with high influence parameters $\alpha_{nc}$ by the proposed method on CIFAR10 data.}
\label{fig:cifar10}
\end{figure*}

\begin{figure*}[t!]
\centering
{\tabcolsep=1em
\begin{tabular}{ccc}
\includegraphics[width=4.3em]{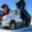}
\includegraphics[width=4.3em]{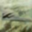}
&
\includegraphics[width=4.3em]{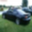}&
\includegraphics[width=4.3em]{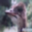}  
\includegraphics[width=4.3em]{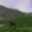}
\includegraphics[width=4.3em]{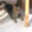}\\
(a) Airplain & (b) Automobile & (c) Bird \\ 
\end{tabular}}
{\tabcolsep=1em
\begin{tabular}{ccccccc}
&
\includegraphics[width=4.3em]{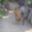}
&
&
\includegraphics[width=4.3em]{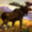}
&
&
\includegraphics[width=4.3em]{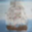}
&
\includegraphics[width=4.3em]{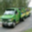}\\
(d) Cat & (e) Deer & (f) Dog & (g) Frog & (h) Horse & (i) Ship & (j) Truck \\
\end{tabular}}
\caption{Ten examples used for the explanations with high influence parameters $\alpha_{nc}$ by the Joint method on CIFAR10 data.}
\label{fig:cifar10joint}
\end{figure*}

\begin{figure*}[t!]
\centering
\begin{tabular}{cccc}
\includegraphics[width=10em]{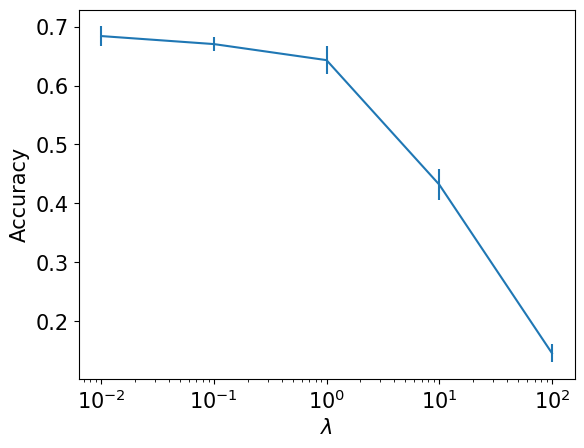}&
\includegraphics[width=10em]{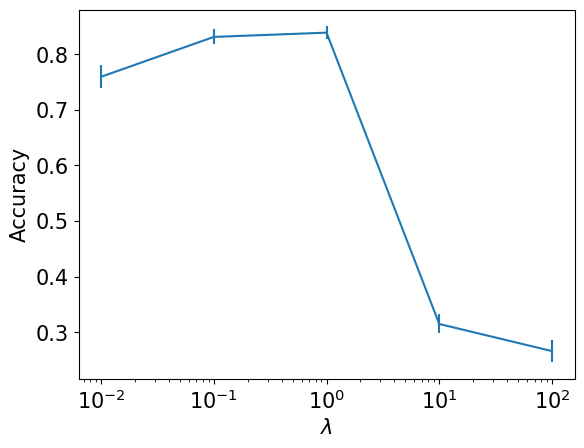}&
\includegraphics[width=10em]{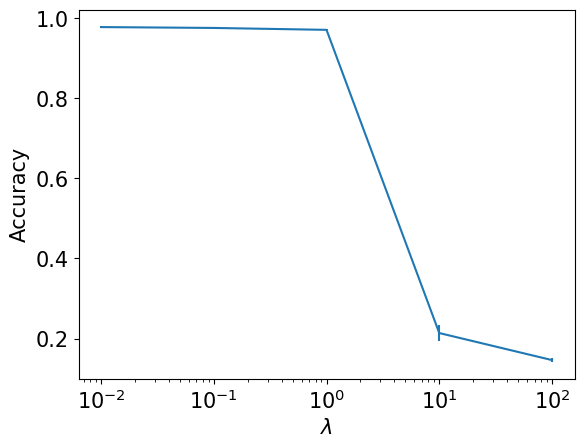}&
\includegraphics[width=10em]{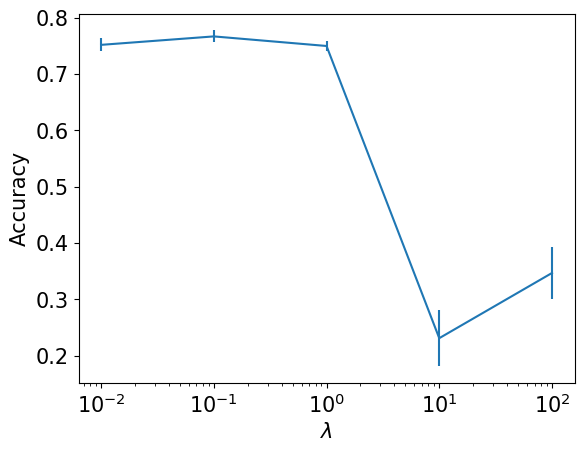}\\
(a) Glass & (b) Vehicle & (c) Segment & (d) CIFAR10 \\
\end{tabular}
\caption{Test accuracy with different $\lambda$. The bar shows the standard error.}
\label{fig:lambda_accuracy}
\end{figure*}

\begin{figure*}[t!]
\centering
\begin{tabular}{cccc}
\includegraphics[width=10em]{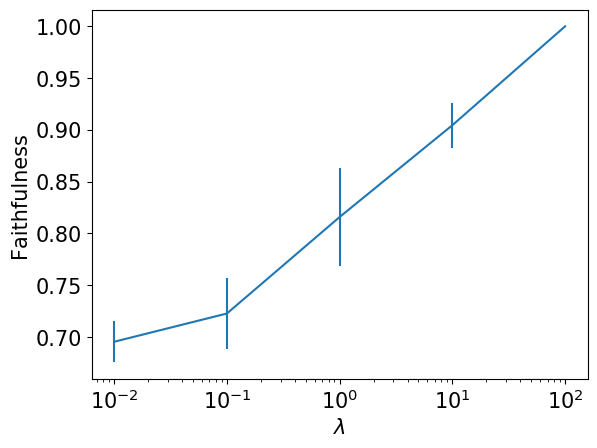}&
\includegraphics[width=10em]{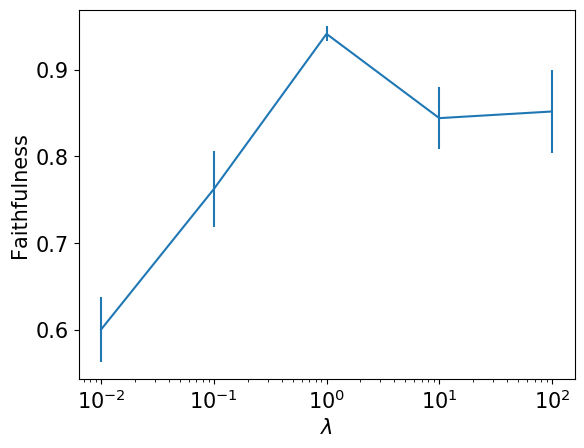}&
\includegraphics[width=10em]{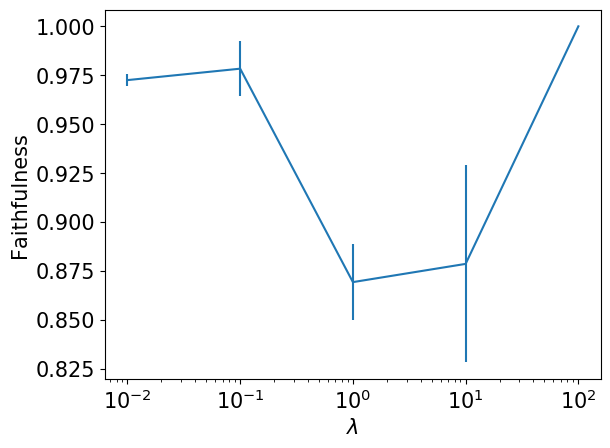}&
\includegraphics[width=10em]{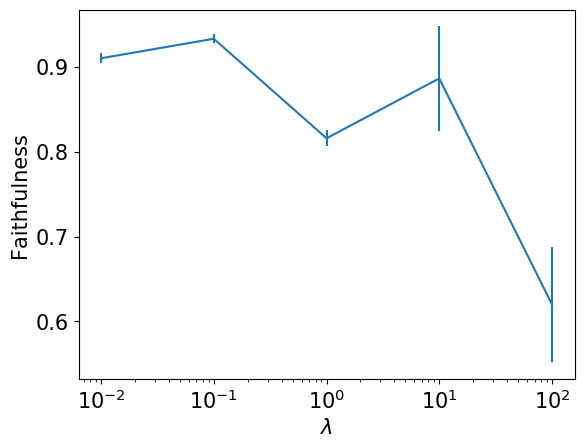}\\
(a) Glass & (b) Vehicle & (c) Segment & (d) CIFAR10 \\
\end{tabular}
\caption{Test faithfulness with different $\lambda$. The bar shows the standard error.}
\label{fig:lambda_faithfulness}
\end{figure*}

\begin{figure*}[t!]
\centering
\begin{tabular}{cccc}
\includegraphics[width=10em]{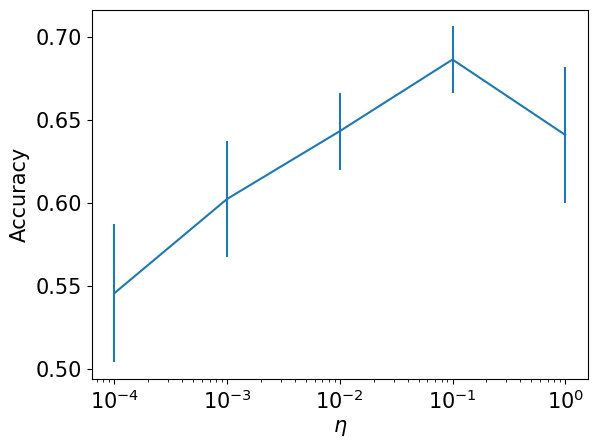}&
\includegraphics[width=10em]{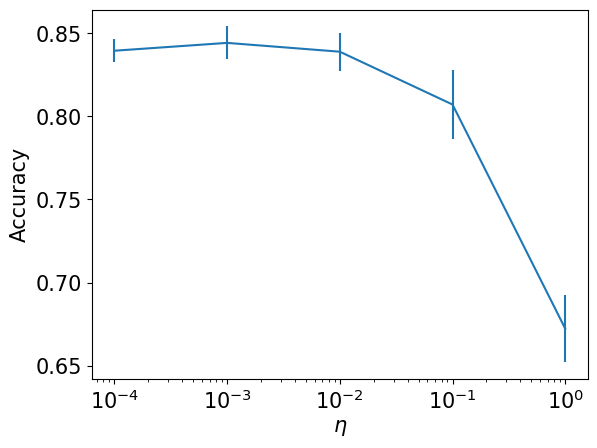}&
\includegraphics[width=10em]{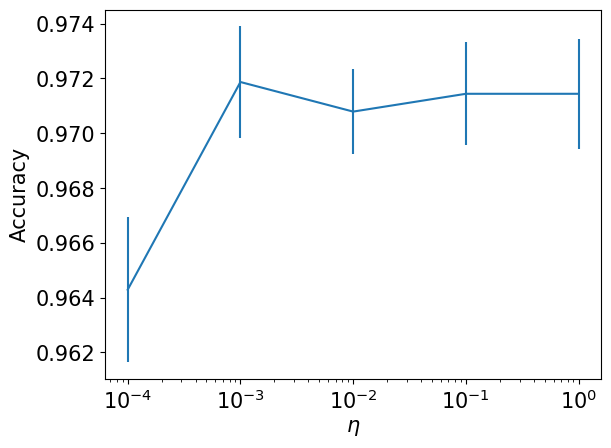}&
\includegraphics[width=10em]{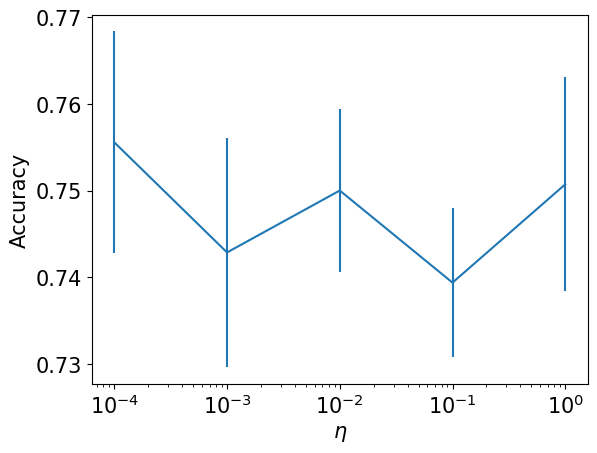}\\
(a) Glass & (b) Vehicle & (c) Segment & (d) CIFAR10 \\
\end{tabular}
\caption{Test accuracy with different $\eta$. The bar shows the standard error.}
\label{fig:eta_accuracy}
\end{figure*}

\begin{figure*}[t!]
\centering
\begin{tabular}{cccc}
\includegraphics[width=10em]{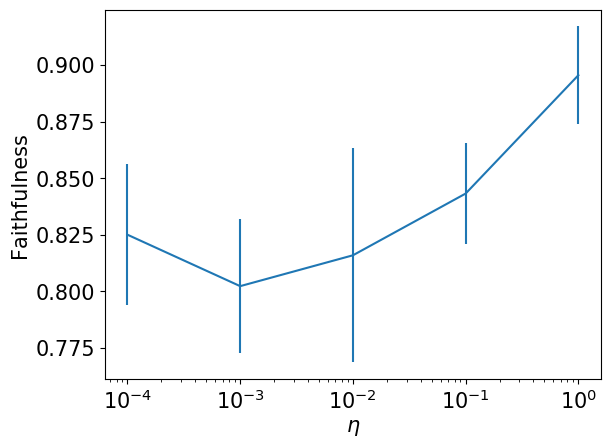}&
\includegraphics[width=10em]{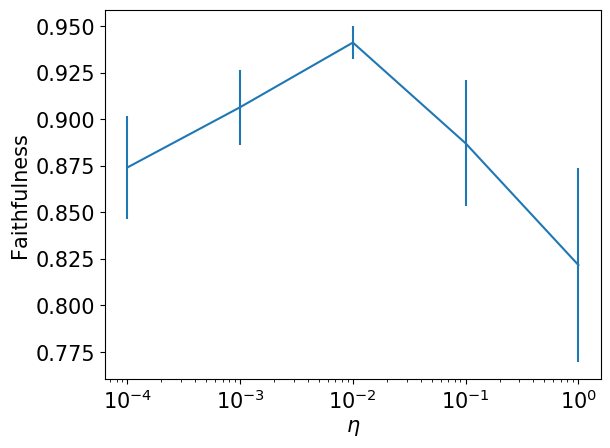}&
\includegraphics[width=10em]{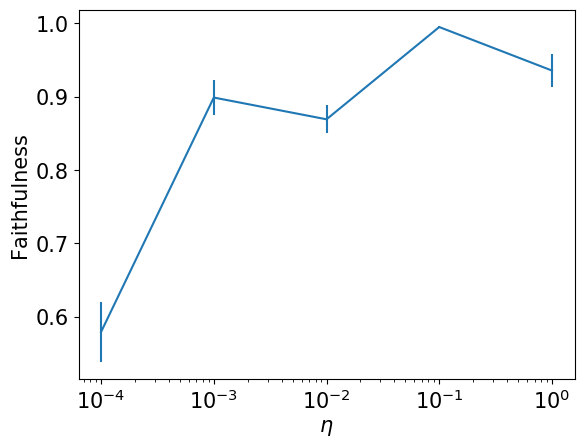}&
\includegraphics[width=10em]{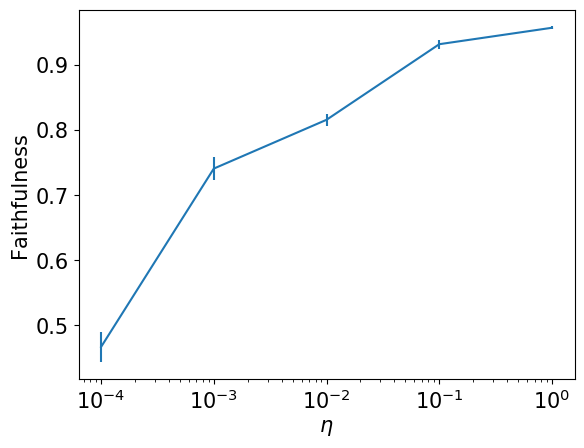}\\
(a) Glass & (b) Vehicle & (c) Segment & (d) CIFAR10 \\
\end{tabular}
\caption{Test faithfulness with different $\eta$. The bar shows the standard error.}
\label{fig:eta_faithfulness}
\end{figure*}

\begin{table*}[t!]
\centering
\caption{Computation time for training in seconds. With Pretrain(R) methods, the time to pretrain the prediction model is not included.}
\label{tab:time}
\begin{tabular}{lrrrrrr}
\hline
&Ours & Joint & RPS & RPSR & Pretrain & PretrainR \\
\hline
Glass & 
635 & 663 & 573 & 543 & 541 & 540 \\
Vehicle & 
2849 & 2628 & 2977 & 2979 & 1800 & 1845 \\
Segment & 
19184 & 19497 & 18731 & 19507 & 10713 & 11330 \\
CIFAR10 & 
2450 & 2422 & 1823 & 1680 & 1769 & 1749 \\
\hline
\end{tabular}
\end{table*}

Figures~\ref{fig:accuracy} and \ref{fig:faithfulness} show
the test accuracy and faithfulness with different numbers of training examples
used in explanation model $h$.
The Pretrain and PretrainR methods achieved the high accuracy
since prediction model $g$ was pretrained without considering explanations by fixing $f$ for training explanation model $h$.
However, their faithfulness was low since it was difficult to approximate prediction model $g$ by
explanation model $h$ with fixed $f$ with a small number of examples.
The faithfulness of the RPS and RPSR methods was always one
since they used explanation model $h$ for prediction.
However, their test accuracy was low since RPS with a small number of examples has a
limited expressive power.
On the other hand, by using different models for predictions and explanations,
the proposed method achieved the test accuracy that was comparable with the pretrain methods,
and the high faithfulness.
This result indicates that the joint training of prediction model $g$
and explanation model $h$ is important.
The accuracy and faithfulness of the Joint method were worse than the proposed method
since the Joint method did not have the regularizer for the sparse explanations.
The accuracy of the RPSR method was better than that of the RPS method,
and the faithfulness of the PretrainR method was better than that of the Pretrain method.
These results indicate the effectiveness of the regularizer
with the stochastic gates in Eq.~(\ref{eq:omega}).

Figures~\ref{fig:cifar10} and \ref{fig:cifar10joint}
show ten training examples used for explanations with high influence parameters $\alpha_{nc}$
on CIFAR10 data by the proposed method and the Joint method, respectively.
With the proposed method, a representative image for each class was selected.
In contrast, the Joint method selected multiple images from a class,
and some classes had no selected examples.
This result indicates that the regularizer is effective to select training examples
appropriately.

Figures~\ref{fig:lambda_accuracy} and \ref{fig:lambda_faithfulness} show
the accuracy and faithfulness with different hyperparameter $\lambda$
by the proposed method with ten examples for explanations.
As $\lambda$ increased, the accuracy decreased, and the faithfulness increased in general.
This result is reasonable since low $\lambda$ values put a high value on improving the accuracy,
which is represented by the first term of the objective function in Eq.~(\ref{eq:E}),
and high $\lambda$ values put a high value on improving the faithfulness,
which is represented by its second term.
With CIFAR10 data, the faithfulness was high with high $\lambda$ values
because the training was unstable. 
Figures~\ref{fig:eta_accuracy} and \ref{fig:eta_faithfulness} show
the accuracy and faithfulness with different regularization hyperparameter $\eta$
by the proposed method with ten examples for explanations.
The best $\eta$ depends on the data, and $\eta$ needs to be tuned by the validation data.

Table~\ref{tab:time} show the computational time for training by the proposed method
We used computers with 2.60GHz CPUs for Glass, Vehicle, and Segment data,
and used computers with 2.30GHz CPU and GeForce GTX 1080 Ti GPU for CIFAR10 data.
The training time of the proposed method was longer than
the RPS(R) and Pretrain(R) methods since the proposed method
needed to train both the prediction and explanation models.
With the explanation model, the number of parameters increases as
the number of the training examples.
Therefore, when there were many training examples,
RPS(R) methods took a longer time than Pretrain(R).
The difference between the computational time with regularization
and that without regularization was small.

\section{Conclusion}
\label{sec:conclusion}

We proposed a method for simultaneously training prediction and explanation models
such that they give faithful example-based explanations with a small number of examples.
In the experiments, we demonstrate that
the proposed method can improve the faithfulness of
the explanation while maintaining the predictive performance.
For future work, we want to apply our framework
to other types of explanation methods,
such as SHAP~\cite{NIPS2017_8a20a862} and
Anchors~\cite{ribeiro2018anchors}.

\bibliographystyle{abbrv}
\bibliography{arxiv_point_selection}

\end{document}